%% file: root.tex
\documentclass[letterpaper, 10 pt, conference]{ieeeconf}  % Comment this line out if you need a4paper

\IEEEoverridecommandlockouts                              % This command is only needed if 
\usepackage[T1]{fontenc}
\usepackage{graphicx}
\usepackage{amsmath}
\usepackage{booktabs}
\usepackage{multirow}
\usepackage{cite}
\usepackage{xcolor}
\usepackage{url}

\newcommand{\yes}{\textcolor{green!55!black}{$\surd$}}
\newcommand{\no}{\textcolor{red!75!black}{$\times$}}
\newcommand{\partly}{\textcolor{orange!85!black}{$\triangle$}}
\newcommand{\rev}[1]{#1}

\title{\LARGE\bfseries
GraspBench: Execution-Aware Cross-Robot Evaluation of 6-DoF Grasp Generation in MuJoCo}

\author{Abdul Basit$^{*}$, Bibek Poudel$^{*}$, and Muhammad Shafique%
\thanks{$^{*}$Equal contribution. The authors are with the eBRAIN Lab,
        Division of Engineering, New York University Abu Dhabi, Abu Dhabi, UAE.
        Email: {\tt \{abdul.basit,bp2376,muhammad.shafique\}@nyu.edu}.}%
}

\begin{document}

\maketitle
\thispagestyle{empty}
\pagestyle{empty}

\input{sections/0-abstract}
\input{sections/1-introduction}
\input{sections/2-background}
\input{sections/3-methods}
\input{sections/4-experiment}
\input{sections/5-results}
\input{sections/6-conclusion}

\bibliographystyle{IEEEtran}
\bibliography{cite}
%%%%%%%%%%%%%%%%%%%%%%%%%%%%%%%%%%%%%%%%%%%%%%%%%%%%%%%%%%%%%%%%%%%%%%%%%%%%%%%%

\end{document}

%% file: sections/0-abstract.tex
\begin{abstract}
\bfseries
\rev{Modern 6-DoF grasp generators are often evaluated using grasp scores, dataset-level metrics, or single-robot demonstrations, yet proposal quality alone does not determine whether a grasp is executable by a particular robot. We present GraspBench, a controlled MuJoCo protocol for evaluating complete grasp-generation pipelines after robot-specific inverse-kinematics filtering, motion planning, and fixed approach-close-lift execution. GraspBench evaluates six configured pipelines across four robot embodiments, five diagnostic objects, and a structured tabletop pose grid, yielding 540 scenarios per method and 3,240 method-scenario evaluations in total. Missing outputs are conservatively counted as failures, and success follows the binary lift criterion produced by each method's execution runner. GraspGen achieves the highest overall success rate (73.1\%), followed by EconomicGrasp (63.5\%), while robot- and object-conditioned analyses reveal substantial variation across embodiments and object geometries. GPG produces the largest native candidate sets but achieves 44.3\% success, indicating that proposal volume alone does not guarantee robot-executable grasps. Cluster-aware paired intervals support the principal method differences while not estimating repeated-execution uncertainty. The results characterize complete configured pipelines on fixed scenarios rather than intrinsic model quality or stochastic repeatability, showing that grasping performance depends jointly on the generator, robot embodiment, scene, and execution protocol.}
\end{abstract}

%% file: sections/1-introduction.tex
\section{Introduction}

Reliable robotic grasping remains a central requirement for practical manipulation. Recent 6-DoF grasp generators can produce dense or diverse candidate grasps from point clouds, and large-scale datasets have accelerated progress in grasp prediction \cite{fang2020graspnet,eppner2021acronym}. Yet a high-scoring grasp proposal is not necessarily executable by a specific robot. The final outcome depends on how the predicted grasp distribution interacts with object geometry, workspace placement, robot kinematics, approach feasibility, collision constraints, and the execution controller. \rev{A well-ranked proposal may be unreachable for one arm, feasible for another, or invalid during approach and lift.}

\rev{This issue becomes sharper across robot embodiments. Differences in reach, joint limits, wrist geometry, and gripper placement can change which candidates survive feasibility filtering. Aggregate success can therefore hide structured failures: a method may perform well on box-like shapes but degrade on thin or concave geometries, succeed near the table center but fail near workspace boundaries, or generate many candidates that downstream constraints discard. These observations motivate evaluating grasp generation as part of an execution pipeline rather than as pose prediction alone.}

Existing grasp resources provide indispensable scale and standardization. GraspNet-1Billion offers dense RGB-D grasp annotations and benchmark metrics \cite{fang2020graspnet}; ACRONYM provides millions of simulated grasps over thousands of objects \cite{eppner2021acronym}; and Dex-Net demonstrated the value of synthetic grasp analysis for robust manipulation policies \cite{mahler2019dexnet}. In parallel, grasp generators such as \rev{GPG, the geometric candidate generator associated with GPD} \cite{tenpas2017gpd}, Contact-GraspNet \cite{sundermeyer2021contact}, AnyGrasp \cite{fang2022anygrasp}, EconomicGrasp \cite{wu2024economicgrasp}, GraspGen \cite{graspgen2025}, and HGGD \cite{chen2024hggd} have advanced geometric, contact-based, dense, budget-aware, generative, and heatmap-guided grasp proposal mechanisms. These works address important parts of the grasping problem, but dataset-level annotations and method-specific demonstrations do not fully answer a complementary question: \textit{how do different generators behave when their outputs are routed through a common scenario-indexed feasibility and simulated-execution structure?} Unlike dataset-level grasp benchmarks that evaluate pose annotations or generator outputs independently of a fixed execution stack, GraspBench fixes the robot, object, workspace, and execution protocol so that generated grasps can be compared under robot-conditioned feasibility constraints.

We study this question through four research questions:
\begin{itemize}
    \item How much do 6-DoF grasp-generator rankings change across robot embodiments?
    \item Which object, workspace, and pose conditions expose failures hidden by aggregate success?
    \item \rev{How do method-native candidate-set sizes relate descriptively to simulated execution success?}
    \item How does failure-aware accounting affect denominator consistency and reported aggregate performance when methods fail to return outputs?
\end{itemize}

\rev{GraspBench evaluates a shared robot-object-workspace-pose grid, enabling paired comparisons across AUBO i5, Franka, KUKA iiwa, and PiPER; five diagnostic objects; workspace positions; and object pitch. For each scenario, a generator proposes grasps, robot-specific filtering selects the highest-ranked feasible candidate, the motion planner produces the robot trajectory, and MuJoCo executes a fixed approach-close-lift sequence scored by its recorded binary lift outcome.}

\rev{The evaluated pipelines span dense perception (AnyGrasp), contact-based prediction (Contact-GraspNet), geometric sampling (GPG), budget-aware generation (EconomicGrasp), generative modeling (GraspGen), and heatmap-guided detection (HGGD). We evaluate execution consequences under scenario-aligned downstream stages, not training recipes or implementation efficiency.}

\rev{Conservative scenario-level accounting counts missing outputs as failures instead of removing them. This prevents missing scenarios or repeated rows from changing the common denominator. Normalized records retain available candidate counts, scores, condition identities, and failure labels for reproducible scenario-level analysis.}

\rev{Three observations emerge. First, method gaps and rankings depend on robot embodiment and object geometry. Second, native proposal volume has no apparent monotonic relation with execution success, although differing budgets preclude a causal candidate-count comparison. Third, missing and explicit no-grasp outputs are execution-relevant outcomes; counting them as failures provides a conservative common denominator rather than excluding absent records.}

The main contributions of this paper are:
\begin{itemize}
    \item \textbf{Execution protocol:} a controlled MuJoCo benchmark protocol for evaluating 6-DoF grasp generators across multiple robot embodiments under a \rev{scenario-aligned} IK-filtered execution pipeline.
    \item \textbf{Scenario diagnostics:} a structured robot-object-workspace-pose grid and paired scenario-level analysis that reveal robot-, object-, workspace-, and pose-conditioned \rev{failure patterns}.
    \item \textbf{Failure-aware study:} \rev{a conservative evaluation of six configured grasp-generation pipelines showing that aggregate and condition-specific execution outcomes can diverge.}
\end{itemize}

\rev{The study is intentionally limited to controlled, single-object simulation. Its five geometries are diagnostic rather than representative of broad object diversity, and the opaque glass model does not test transparent-object perception. GraspBench therefore complements, rather than replaces, cluttered-scene benchmarks, candidate-level Success@K analysis, repeated stochastic trials, and real-robot validation.}

%% file: sections/2-background.tex
\section{Background and Related Work}

\subsection{6-DoF Grasp Generation and Datasets}

Large-scale grasp datasets and simulation pipelines have played a major role in modern grasping. Dex-Net demonstrated that synthetic grasp analysis can train robust grasping policies \cite{mahler2019dexnet}, and the YCB object set helped standardize manipulation objects and models \cite{calli2015ycb}. GraspNet-1Billion introduced a large-scale RGB-D benchmark for general object grasping \cite{fang2020graspnet}, while ACRONYM provided millions of simulated grasps over thousands of objects \cite{eppner2021acronym}. Together, these resources provide broad object coverage, large-scale data, and standardized candidate evaluation.

Several families of 6-DoF grasp generators are relevant to our study. GPD estimates grasp poses directly from point clouds using local geometric evidence \cite{tenpas2017gpd}, and PointNetGPD learns grasp scoring from point sets \cite{liang2019pointnetgpd}. Contact-GraspNet predicts contact-based 6-DoF grasps from cluttered scenes \cite{sundermeyer2021contact}. AnyGrasp emphasizes robust and efficient grasp perception across spatial and temporal domains \cite{fang2022anygrasp}. EconomicGrasp studies efficient candidate generation under a 6-DoF detection framework \cite{wu2024economicgrasp}, while HGGD uses heatmap-guided grasp detection \cite{chen2024hggd}. Generative approaches, including 6-DOF GraspNet \cite{mousavian20196dofgraspnet} and GraspGen \cite{graspgen2025}, model distributions over grasp poses and can produce diverse hypotheses. \rev{Their distinct proposals motivate robot-specific, complete-pipeline evaluation.}

GraspBench is also related to broader simulation benchmarks for manipulation. Robosuite provides modular MuJoCo environments for robot-learning evaluation \cite{zhu2020robosuite}, RLBench defines a large set of language-conditioned manipulation tasks \cite{james2020rlbench}, Meta-World studies multi-task and meta-reinforcement-learning manipulation \cite{yu2020metaworld}, and ManiSkill and SAPIEN provide scalable simulated manipulation environments and assets \cite{mu2021maniskill,xiang2020sapien}. These benchmarks emphasize policy learning, environment diversity, or general manipulation tasks. \rev{GraspBench instead diagnoses generators under scenario-aligned tabletop feasibility and execution.}

\begin{table}[ht]
\centering
\caption{Evaluation-scope comparison with representative grasping datasets, generators, and manipulation benchmarks.}
\label{tab:related_positioning}
\scriptsize
\setlength{\tabcolsep}{2.0pt}
\resizebox{0.98\columnwidth}{!}{%
\begin{tabular}{@{}llccccc@{}}
\toprule
\textbf{Work} & \textbf{Role} & \textbf{Scale} & \textbf{Exec.} & \textbf{$\times$-robot} & \textbf{Cond.} & \textbf{Fail.}\\
\midrule
Dex-Net~\cite{mahler2019dexnet} & synthetic policy/data & \yes & \yes & \no & \partly & \no\\
GraspNet~\cite{fang2020graspnet} & RGB-D benchmark & \yes & \partly & \no & \partly & \no\\
ACRONYM~\cite{eppner2021acronym} & simulated grasp data & \yes & \partly & \no & \no & \no\\
Robosuite~\cite{zhu2020robosuite} & manipulation sim & \partly & \yes & \partly & \yes & \no\\
RLBench~\cite{james2020rlbench} & task benchmark & \yes & \yes & \partly & \yes & \no\\
\rev{GPD/GPG}~\cite{tenpas2017gpd} & \rev{geom. generator} & \no & \partly & \no & \partly & \no\\
\rev{CGN}~\cite{sundermeyer2021contact} & contact generator & \partly & \partly & \no & \partly & \no\\
AnyGrasp~\cite{fang2022anygrasp} & dense perception & \partly & \yes & \no & \partly & \no\\
EconomicGrasp~\cite{wu2024economicgrasp} & budget-aware detector & \partly & \partly & \no & \partly & \no\\
GraspGen~\cite{graspgen2025} & generative model & \partly & \partly & \partly & \partly & \no\\
HGGD~\cite{chen2024hggd} & heatmap-guided detector & \partly & \partly & \no & \partly & \no\\
\textbf{GraspBench} & execution diagnostic protocol & \partly & \yes & \yes & \yes & \yes\\
\bottomrule
\end{tabular}
}
\parbox{0.98\columnwidth}{\vspace{0.6mm}\scriptsize Scale=large benchmark; Exec.=robot/physics execution; $\times$-robot=cross-embodiment; Cond.=conditioned analysis; Fail.=failure accounting. Symbols: \yes~=core, \partly~=partial/indirect, \no~=not primary (not necessarily absent).}
\end{table}

\begin{figure*}[ht]
    \centering
    \includegraphics[width=\textwidth]{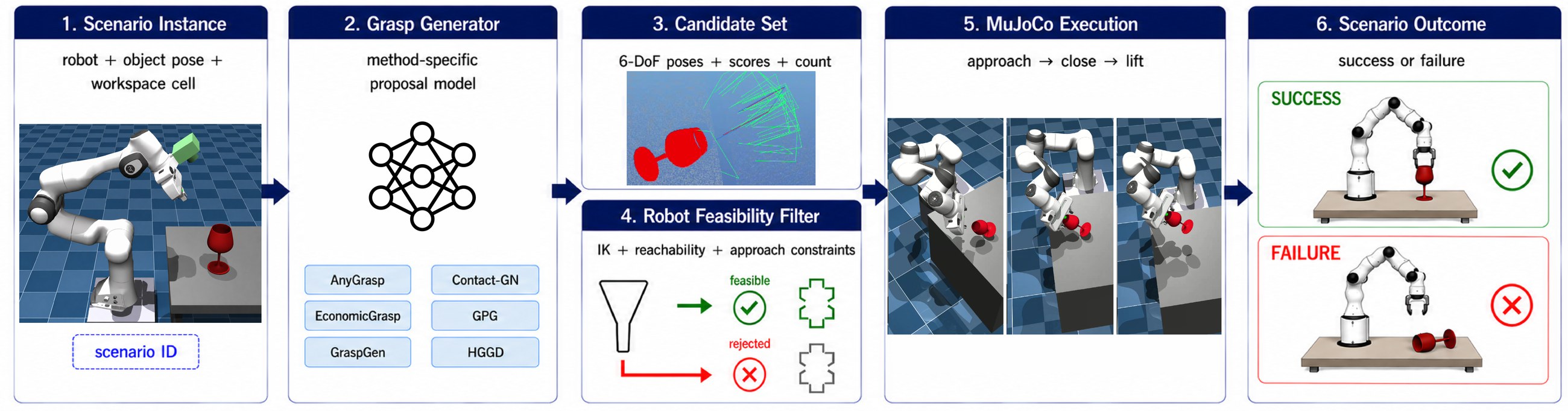}
    \caption{\rev{GraspBench pipeline: proposal, robot-specific feasibility filtering, OMPL planning, MuJoCo execution, and outcome logging. Missing outputs and scenarios with no feasible grasp count as failures; Contact-GN denotes Contact-GraspNet (CGN).}}
    \label{fig:pipeline}
\end{figure*}

Table~\ref{tab:related_positioning} summarizes the intended positioning of GraspBench relative to these resources and representative grasp-generation methods. While large-scale datasets emphasize object coverage and standardized grasp scoring, and method-specific studies evaluate individual generators and, in some cases, real-robot performance, GraspBench focuses on controlled execution-level diagnosis across multiple robot embodiments, object poses, and workspace conditions, with explicit failure accounting. \rev{This execution-focused evaluation complements large-scale grasp benchmarks and real-robot validation.}

\subsection{Execution-Aware Evaluation}

\rev{Robot execution introduces constraints that grasp scores alone do not capture. Candidate poses must satisfy reachability, IK feasibility, collision avoidance, motion planning, gripper closure, and successful lifting under the recorded outcome criterion. MuJoCo provides a controlled physics environment for studying these interactions \cite{todorov2012mujoco}. GraspBench evaluates each configured generator after robot-specific feasibility and a fixed high-level tabletop execution sequence. The objective is not to claim real-world transfer, but to reveal where proposal rankings and robot-executable outcomes diverge.}

%% file: sections/3-methods.tex
\section{\rev{GraspBench Design}}

\subsection{Benchmark Objective}

GraspBench is designed to answer a system-level question: \textit{given a grasp generator, how often does it produce an executable grasp for a specific robot, object, and tabletop configuration?} To isolate this question, all main results are evaluated under an IK-filtered execution protocol. The executed grasp is the highest-ranked feasible candidate selected by the benchmark pipeline from the method's scored or ordered proposals. \rev{Candidate ordering is method-native: scores determine order only within a pipeline and score magnitudes are never compared across methods. No common candidate budget or oracle Success@K analysis is claimed.} A feasible candidate is one accepted by the benchmark's robot-specific IK, reachability, and approach-constraint selection stage. The normalized records expose scenario outcomes and, where available, tested IK candidate counts; the reported metrics operate at the scenario level. A scenario is counted as successful only if the selected grasp completes the fixed MuJoCo approach-close-lift trial and \rev{the corresponding execution runner records a successful binary lift outcome}. If a method fails to return an output for a scenario, the scenario is counted as a failure. \rev{Figure~\ref{fig:pipeline} summarizes the stages.}

\subsection{Scenario Grid}

Table~\ref{tab:grid} summarizes the benchmark grid. Each method is evaluated over four robots, five diagnostic objects, three distance bands, three lateral positions, and three pitch values, giving 540 intended scenarios. \rev{In MuJoCo world coordinates, Near, Mid, and Far correspond to $x=0.45$, $0.55$, and $0.65$~m; Left, Center, and Right correspond to $y=0.10$, $0$, and $-0.10$~m; pitch is applied about the object's $x$-axis. Object origins use $z=0.55$~m ($0.555$~m for the cylinder).} The robots are AUBO i5, \rev{Franka Emika Panda, KUKA LBR iiwa 14, and AgileX PiPER}, which provide different arm geometries, reachable workspaces, wrist configurations, and simulated end-effectors. The object set contains an airplane model, glass, cube, cylinder, and bottle, spanning simple convex shapes, elongated shapes, and thin or concave geometries. The distance, lateral-position, and pitch grid is designed to probe reachability, workspace-boundary effects, and orientation sensitivity. \rev{Because grippers, camera mounts, controllers, grasp frames, and collision models are robot-specific, cross-robot differences characterize complete embodiment configurations and cannot be attributed to arm reach alone.}

\begin{figure}[ht]
    \centering
    \includegraphics[width=\linewidth]{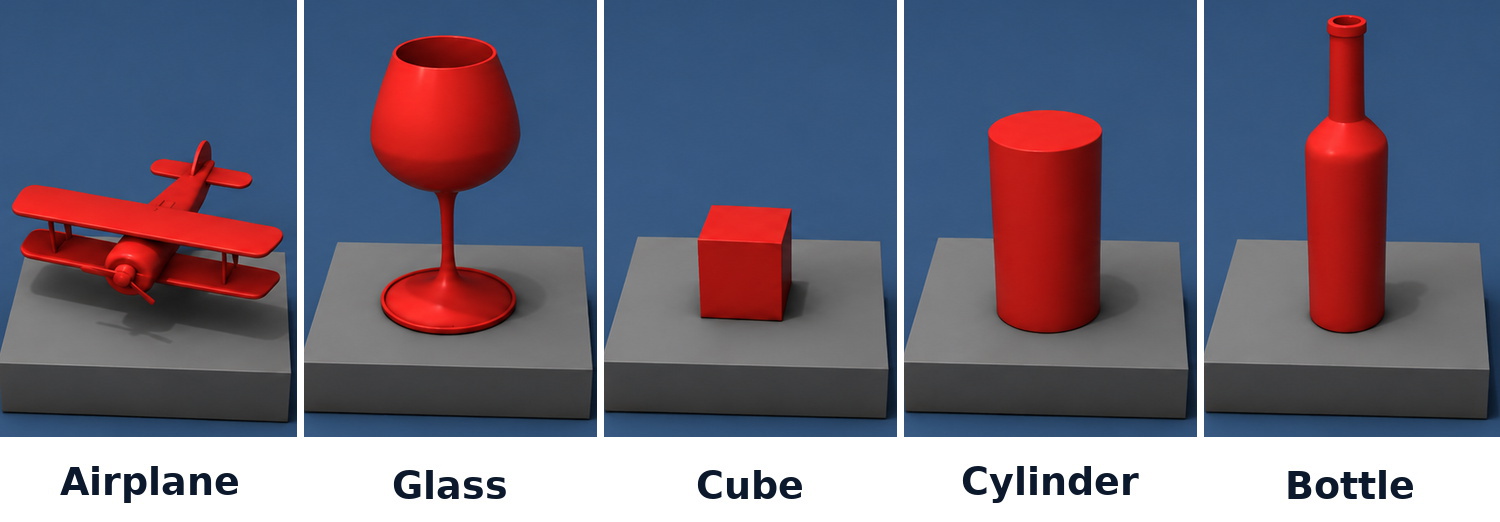}
    \caption{\rev{Diagnostic GraspBench object set.}}
    \label{fig:object_set}
\end{figure}

\begin{figure}[ht]
    \centering
    \includegraphics[width=\linewidth]{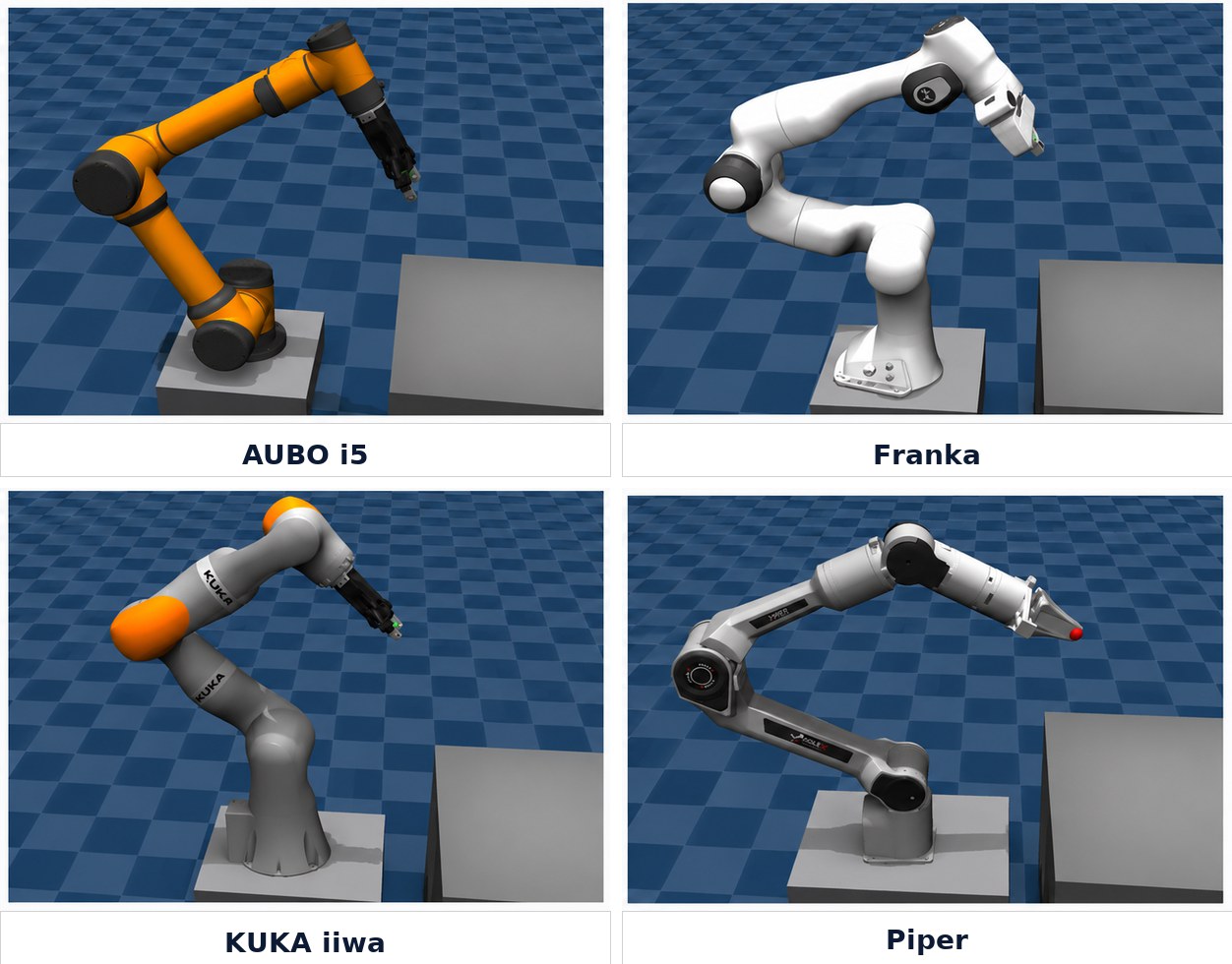}
    \caption{\rev{Evaluated robot configurations; each includes its arm, gripper, camera, controller, frame adapter, and collision model. ``Piper'' denotes AgileX PiPER.}}
    \label{fig:robot_set}
\end{figure}

\begin{figure}[ht]
    \centering
    \includegraphics[width=\linewidth]{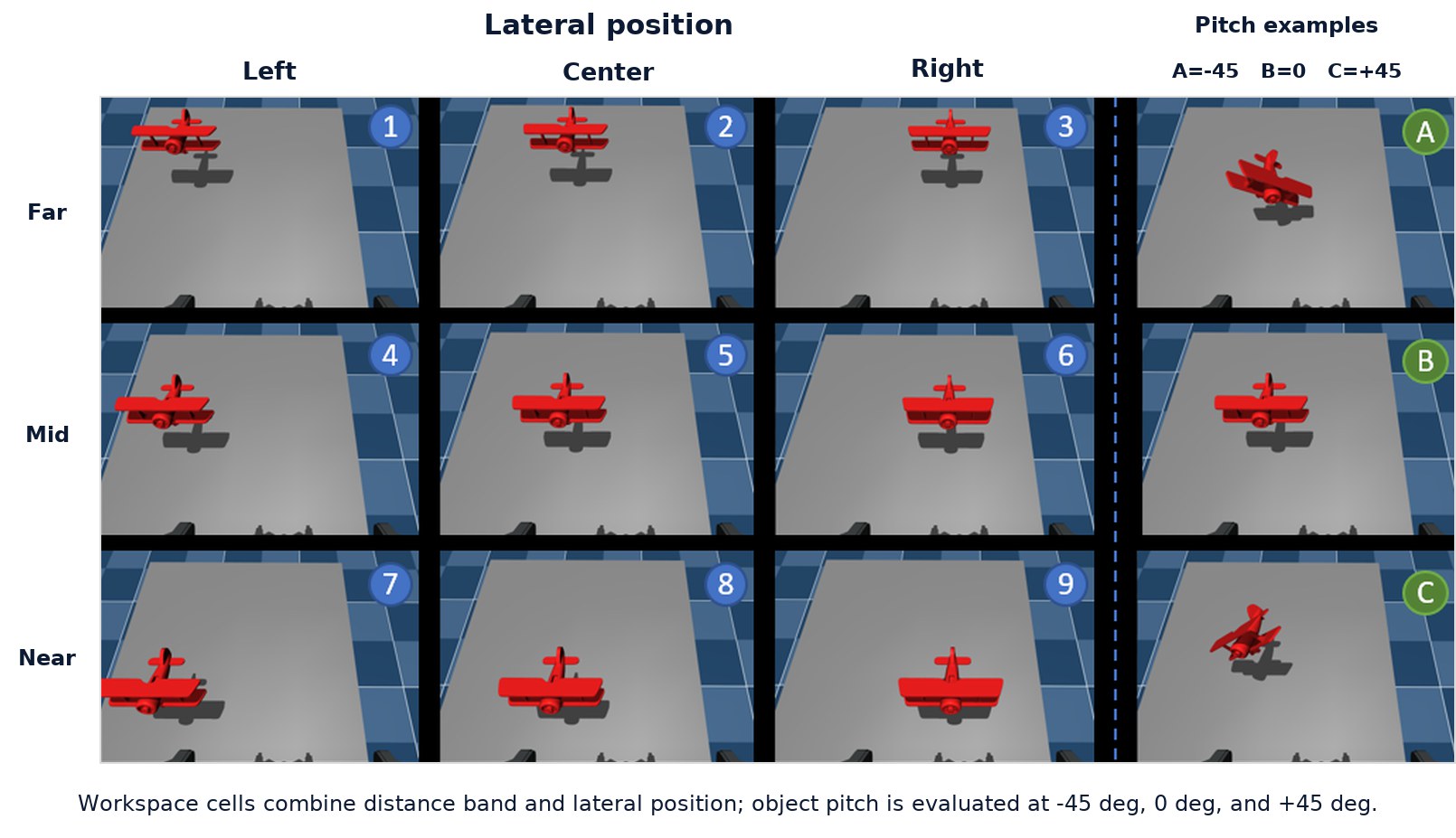}
    \caption{\rev{Tabletop distance, lateral-position, and pitch grid for condition-specific analysis.}}
    \label{fig:workspace_grid}
\end{figure}

\rev{Figures~\ref{fig:object_set}, \ref{fig:robot_set}, and~\ref{fig:workspace_grid} visualize the benchmark factors.}

\begin{table}[ht]
\centering
\caption{GraspBench scenario grid used in the main evaluation.}
\label{tab:grid}
\scriptsize
\setlength{\tabcolsep}{3.0pt}
\resizebox{\columnwidth}{!}{%
\begin{tabular}{@{}llr@{}}
\toprule
\textbf{Category} & \textbf{Values} & \textbf{Count} \\
\midrule
Robots & AUBO i5, Franka, KUKA iiwa, \rev{PiPER} & 4 \\
Methods & AG, CGN, Econ., GPG, GG, HGGD & 6 \\
Objects & airplane, glass, cube, cylinder, bottle & 5 \\
Distance & Near, Mid, Far & 3 \\
Lateral & Left, Center, Right & 3 \\
Pitch & $-45^\circ$, $0^\circ$, $45^\circ$ & 3 \\
\midrule
Scenarios/method & Full grid & 540 \\
Method-scenarios & $540 \times 6$ methods & 3240 \\
\bottomrule
\end{tabular}
}

\parbox{0.98\columnwidth}{\vspace{0.6mm} \scriptsize Method abbreviations: AG = AnyGrasp, CGN = Contact-GraspNet, Econ. = EconomicGrasp, and GG = GraspGen.}
\end{table}

\subsection{Methods}

We evaluate six grasp-generation pipelines: AnyGrasp (AG), Contact-GraspNet (CGN), EconomicGrasp (Econ.), \rev{GPG, the geometric candidate generator associated with GPD}, GraspGen (GG), and HGGD. These methods span dense detection, contact-based prediction, geometric candidate generation, budget-aware/fixed-budget sampling, generative grasp modeling, and heatmap-guided detection. The analysis focuses on the evaluation layer: method outputs retain their native within-method ordering, are mapped to the same scenario keys, and pass through the same failure-aware accounting. \rev{Accordingly, the results compare configured end-to-end pipelines under a scenario-aligned IK-filtered execution structure rather than training procedures, hardware-normalized runtime, or model architectures in isolation.}

\subsection{Failure-Aware Data Treatment}

\rev{We use three failure terms consistently: a \emph{missing output} has no raw scenario record, a \emph{no-grasp output} explicitly records an empty/no-grasp result, and an \emph{other observed failure} records unsuccessful execution without a recoverable first-failure stage.} Missing entries indicate scenarios for which no method output is available; because such entries cannot be verified as successful executions, they are conservatively counted as failures. This yields an equal intended denominator of 540 scenarios for every method. Duplicate observed rows are collapsed at the scenario level to avoid overweighting repeated entries. This policy prevents optimistic success estimates that could arise if missing scenarios were excluded from the denominator, particularly when missingness is non-random, and makes the comparison reproducible from the scenario grid.

%% file: sections/4-experiment.tex
\section{Experimental Setup}

\subsection{Simulation Environment}

All experiments are evaluated in a MuJoCo-based tabletop setting. Each trial places one target object at a specified distance band, lateral position, and pitch. A grasp-generation method proposes candidate grasps, the pipeline applies robot-specific feasibility filtering, and the selected feasible grasp passes through motion planning before simulated execution. The primary outcome is binary grasp success after the fixed approach-close-lift execution. Because the accompanying records encode lift success as a binary scenario-level outcome, we evaluate the recorded outcome directly and do not introduce an additional post hoc height or stability threshold.

\rev{Each robot uses one arm-mounted MuJoCo camera with $60^\circ$ vertical field of view; RGB-D frames are back-projected into partial point clouds in the camera frame. The benchmark uses synthetic depth without added sensor or calibration noise and does not test transparent-surface sensing. Camera placement and resulting self-occlusion are embodiment-specific. Method adapters retain their native preprocessing and candidate ordering.}

\rev{The execution wrapper maps camera-relative $4\!\times\!4$ grasp poses into each robot's base/end-effector frames. It uses damped-least-squares IK (up to 2000 iterations), a 0.02~m positional acceptance threshold, joint limits, and MuJoCo contact checks for robot collision. Motion planning uses OMPL RRTConnect~\cite{sucan2012ompl} with a 2~s timeout and 80 interpolated samples for reach and 40 each for approach and lift. Planner choice was not varied as an experimental factor, so its effect is not analyzed separately. AUBO i5 and KUKA iiwa use AG95 end effectors, whereas Franka and PiPER use their model-specific grippers; approach/lift offsets, gains, closure commands, and motion durations are consequently robot-specific rather than controlled arm-only variables. Full per-robot values are specified in the execution implementation.}

\rev{All methods use the same intended scenario grid and the same high-level feasibility, planning, and approach-close-lift stages for a fixed embodiment. Success is analyzed using the binary lift label produced by each execution runner. GraspGen and AnyGrasp automatically mark lifts of at least 0.10~m as successful, with recorded manual judgment for lower lifts; the remaining runners directly record the final binary lift outcome. This preserves each pipeline's implemented success decision while applying common scenario-level accounting.}

The benchmark protocol uses a fixed grid rather than randomized clutter. This choice makes cross-method and cross-robot comparisons interpretable: every method is evaluated against the same intended scenario set, allowing paired scenario-level analysis. \rev{The normalized table contains one row per intended method-scenario key; missing raw method records are explicitly marked and counted as failures. The records do not encode repeated stochastic trials or random seeds. Therefore,} the present study estimates scenario-level behavior rather than repeatability under simulator noise. The current study does not claim robustness to clutter, occlusion, or real-world transfer. \rev{Matched ablations without IK filtering or lift judgment, repeated seeded executions, and cluttered scenes require new candidate-level runs and are outside the present scope.}

\subsection{Analysis Artifacts}

The records provide scenario-level execution outcomes, candidate counts, selected scores where available, tested IK candidates for some methods, and logged pipeline time. Candidate-level trace analysis and finer collision-stage decomposition are planned extensions.

\subsection{Metrics}

The main metric is success rate:
\begin{equation}
SR = \frac{N_{\mathrm{success}}}{N_{\mathrm{scenarios}}},
\end{equation}
where $N_{\mathrm{scenarios}}=540$ for every method in the main comparison. We additionally report robot-conditioned success, object-conditioned success, pitch-conditioned success, \rev{native candidate counts, returned-record availability, success conditional on a returned record,} and failure accounting. Missing outputs are included in the denominator as failures. \rev{Wilson 95\% intervals summarize binary outcomes over the finite scenario grid; they are not confidence intervals for repeated stochastic execution.}

For pairwise method comparisons, we compute per-scenario differences on the common 540-scenario grid. A positive value for method $A-B$ indicates that $A$ succeeded where $B$ failed more often than the reverse. \rev{To account for dependence among poses of the same embodiment and object, 95\% percentile intervals use 50,000 bootstrap resamples of the 20 robot-object clusters; all 27 pose/workspace outcomes within a sampled cluster remain together. These intervals describe sensitivity across scenario groups, not execution repeatability.}

\subsection{\rev{Candidate Records and Runtime Scope}}

\rev{Candidate counts describe method-native outputs before method-specific truncation and feasibility testing; confidence calibration, suppression, and sampling budgets are not standardized. We therefore do not use them for controlled efficiency or Success@K claims. Logged times are not comparable across implementations because hardware, software, and stage boundaries differ.}

% \subsection{Accounting Cases}

% Two accounting cases require explicit tracking: \rev{missing outputs}, counted as failures, and duplicated rows, collapsed before aggregation. \rev{Section~V reports the effect for each method.}

%% file: sections/5-results.tex
\section{Results and Discussion}

\subsection{Overall IK-Filtered Execution Performance}

Table~\ref{tab:main_results} summarizes the main results. \rev{The configured GraspGen pipeline} achieves the highest success rate, with 395 successes out of 540 scenarios (73.1\%). EconomicGrasp \rev{ranks second}, with 343 successes (63.5\%). \rev{AnyGrasp, HGGD, and GPG range from 44.3--48.0\%, versus 33.0\% for Contact-GraspNet.}

\begin{table}[ht]
\centering
\caption{\rev{IK-filtered results. Missing outputs fail; intervals are Wilson grid summaries; candidates are native mean/median.}}
\label{tab:main_results}
\scriptsize
\setlength{\tabcolsep}{2.2pt}
\resizebox{\columnwidth}{!}{%
\begin{tabular}{lccccc}
\toprule
\textbf{Method} & \textbf{Success n/540 (\%)} & \textbf{95\% int. (\%)} & \rev{\textbf{Missing}} & \textbf{Cand. mean/med.} & \rev{\textbf{IK max}} \\
\midrule
GraspGen & 395 (73.1) & [69.3, 76.7] & 0 & 296/304 & \rev{50} \\
EconomicGrasp & 343 (63.5) & [59.4, 67.5] & 0 & 1024/1024 & \rev{50} \\
AnyGrasp & 259 (48.0) & [43.8, 52.2] & 1 & 391/365 & \rev{80} \\
HGGD & 249 (46.1) & [41.9, 50.3] & 5 & 6/6 & \rev{20} \\
GPG & 239 (44.3) & [40.1, 48.5] & 0 & 7414/8088 & \rev{--} \\
Contact-GraspNet & 178 (33.0) & [29.1, 37.0] & 31 & 23/23 & \rev{--} \\
\bottomrule
\end{tabular}
}
\parbox{0.98\columnwidth}{\vspace{0.5mm}\scriptsize \rev{IK max is the maximum tested-candidate count observed in available logs; ``--'' denotes an unrecorded field, not zero tests.}}
\end{table}

\begin{figure}[ht]
    \centering
    \includegraphics[width=\columnwidth]{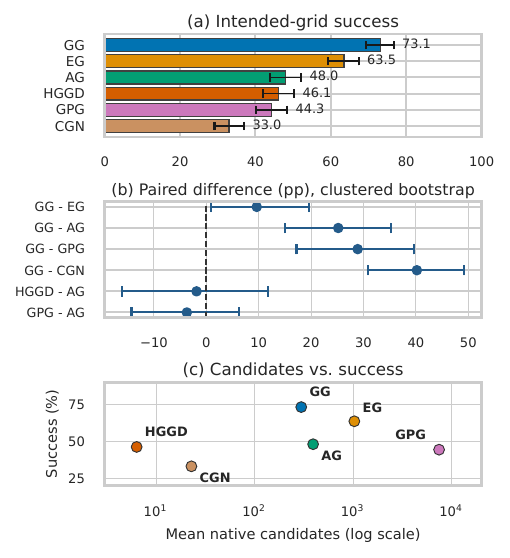}
    \caption{\rev{Execution results: (a) success with Wilson intervals; (b) clustered paired differences; (c) native candidates versus success (descriptive because budgets differ). Abbreviations follow Table~\ref{tab:grid}.}}
    \label{fig:headline}
\end{figure}

\rev{Figure~\ref{fig:headline} shows the intended-grid results and cluster-aware paired analysis. Resampling robot-object groups gives GraspGen paired advantages of 9.6 percentage points over EconomicGrasp (95\% interval 0.9--19.6), 25.2 over AnyGrasp (15.0--35.2), 28.9 over GPG (17.2--39.6), and 40.2 over Contact-GraspNet (30.9--49.3). HGGD--AnyGrasp ($-1.9$, $-16.1$--11.9) and GPG--AnyGrasp ($-3.7$, $-14.3$--6.3) cross zero. These intervals summarize variation across 20 robot-object clusters and do not measure repeated-execution uncertainty.}

\subsection{Cross-Robot Sensitivity}

\begin{figure}[ht]
    \centering
    \includegraphics[width=\columnwidth]{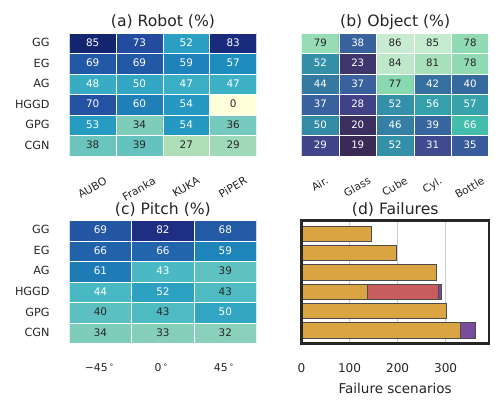}
    \caption{\rev{Diagnostics by (a) robot, (b) object, (c) pitch, and (d) failure type: other observed (amber), no grasp (red), and missing output (purple). Abbreviations follow Table~\ref{tab:grid}.}}
    \label{fig:diagnostics}
\end{figure}

\rev{Figure~\ref{fig:diagnostics}(a) shows substantial variation across robot configurations.} GraspGen reaches 85.2\% on AUBO i5 and 83.0\% on \rev{PiPER}, but drops to 51.9\% on KUKA iiwa. EconomicGrasp is more uniform, ranging from 57.0\% on \rev{PiPER} to 68.9\% on AUBO i5 and Franka. GPG performs similarly on AUBO i5 and KUKA iiwa (53.3\% and 54.1\%) but is lower on Franka and \rev{PiPER}. \rev{HGGD records nonzero success on AUBO i5, Franka, and KUKA iiwa, but zero successes on PiPER. All 135 HGGD/PiPER scenarios are explicitly logged as no-grasp outputs; the records do not localize the cause to generation, frame adaptation, or feasibility filtering. We therefore describe a configuration-level failure pattern, not intrinsic HGGD quality. Robot scale may be one contributing factor, but the available records do not isolate it.} As a sensitivity check, excluding \rev{PiPER} gives HGGD 249/405 successes (61.5\%) across the remaining robots; over the same non-\rev{PiPER} subset, GraspGen reaches 69.9\% and EconomicGrasp reaches 65.7\%, \rev{and these two pipelines remain ahead of HGGD on this subset.}

\rev{This variation characterizes complete embodiment configurations: arm kinematics, end effector, camera and base mounting, grasp-frame adapter, controller, and collision model vary jointly. Isolating the contribution of arm reach requires a controlled component-level ablation.}

\subsection{Object and Pose Sensitivity}

Figure~\ref{fig:diagnostics}(b) shows object-conditioned success. Cube is the easiest object overall, while the glass is the hardest. GraspGen performs strongly on cube, cylinder, bottle, and airplane model, but drops to 38.0\% on the glass. EconomicGrasp is similarly strong on cube (84.3\%), cylinder (80.6\%), and bottle (77.8\%), but reaches only 23.1\% on the glass. Thus, even the strongest methods remain sensitive to thin or difficult geometries. \rev{Qualitatively, some successful glass trials used top- or rim-directed grasps rather than potentially more task-suitable stem or side contacts. Binary lift success therefore need not indicate an optimal or semantically appropriate grasp.}

Pitch-conditioned results in Fig.~\ref{fig:diagnostics}(c) reveal additional structure. GraspGen peaks at zero pitch (81.7\%) and decreases at $\pm45^\circ$. EconomicGrasp remains high at $-45^\circ$ and $0^\circ$ but drops at $45^\circ$. AnyGrasp also decreases at $45^\circ$, while GPG improves at $45^\circ$. Workspace effects are weaker than robot and object effects in this grid: averaged over methods, success is 49.2\% for Near, 52.8\% for Mid, and 52.0\% for Far placements, with lateral averages of 49.4\% for Center, 53.7\% for Left, and 50.9\% for Right. Thus, workspace factors are most useful here as paired scenario controls, while object pose and robot embodiment produce clearer diagnostic variation.

\subsection{Failure Accounting}

Figure~\ref{fig:diagnostics}(d) reports failure accounting. Contact-GraspNet has 31 missing outputs; these are counted as failures and reduce its final success rate to 33.0\%. HGGD has five missing outputs and 148 no-grasp outputs, including all \rev{PiPER} scenarios. AnyGrasp has one missing output. This conservative accounting prevents optimistic estimates that could result from excluding missing scenarios. \rev{The reported labels distinguish missing, no-grasp, and other observed failures; recording the first execution stage would enable finer attribution to IK, collision, planning, closure, drop, or lift.}

\begin{table}[ht]
\centering
\caption{\rev{Returned-record availability and success (\%). End-to-end SR uses all 540 intended scenarios.}}
\label{tab:accounting_sensitivity}
\small
\setlength{\tabcolsep}{2.0pt}
\renewcommand{\arraystretch}{0.96}
\begin{tabular}{@{}lccc@{}}
\toprule
\textbf{Method} & \textbf{Returned} & \textbf{E2E SR} & \textbf{SR\textbar record} \\
 & \textbf{records} & \textbf{(\%)} & \textbf{(\%)} \\
\midrule
GraspGen & 540 (100.0) & 73.1 & 73.1 \\
EconomicGrasp & 540 (100.0) & 63.5 & 63.5 \\
AnyGrasp & 539 (99.8) & 48.0 & 48.1 \\
HGGD & 535 (99.1) & 46.1 & 46.5 \\
GPG & 540 (100.0) & 44.3 & 44.3 \\
Contact-GraspNet & 509 (94.3) & 33.0 & 35.0 \\
\bottomrule
\end{tabular}
\parbox{0.98\columnwidth}{\vspace{0.4mm}\scriptsize \rev{Stage-conditional success after IK or planning is unavailable because those denominators were not recorded consistently.}}
\end{table}

\rev{Table~\ref{tab:accounting_sensitivity} shows that conditioning on a returned record changes success by only 0.1 percentage points for AnyGrasp, 0.4 for HGGD, and 2.0 for Contact-GraspNet. HGGD's dominant recorded issue is therefore the explicit no-grasp output rather than missing records. The unavailable stage denominators prevent stronger attribution to IK, planning, or execution control.}

\subsection{Implications}

The results support several practical conclusions. First, execution-aware benchmarking can change method interpretation: the highest-scoring grasp proposal is not necessarily reachable, IK-feasible, or plannable for a particular robot. \rev{This motivates explicit robot-specific filtering, although its marginal effect is not isolated because the records do not support a matched no-IK ablation. Its cost is also not quantified because timing was not measured on common hardware. Across the configured pipelines, no monotonic relation is apparent between native candidate count and success; differing budgets and filtering preclude causal interpretation.} Partial point-cloud observations can make object extent and grasp pose estimation ambiguous, suggesting that geometric context, camera-view context, or multi-view perception may be needed for difficult objects. Finally, conservative handling of missing outputs and duplicate records is necessary for defensible benchmark reporting. \rev{These findings motivate condition-specific, execution-aware, failure-aware reporting.}

%% file: sections/6-conclusion.tex
\section{Conclusion and Limitations}

\rev{GraspBench provides an execution-aware MuJoCo protocol for six grasp pipelines, four robots, five diagnostic objects, and a structured tabletop grid. Common scenarios, missing-output failures, and duplicate collapse support conservative paired comparisons. The configured GraspGen pipeline leads at 73.1\%, followed by EconomicGrasp at 63.5\%, while robot- and object-conditioned results vary substantially. GPG generates the largest native candidate sets but reaches 44.3\% success, showing that proposal volume alone does not ensure robot-executable grasps; candidate counts remain descriptive because budgets differ. Thus, grasp performance is a systems property of the generator, embodiment, scene, and execution stack. Conclusions apply to complete configured pipelines and one execution per scenario in single-object simulation; isolating individual components requires controlled ablations. Future work includes repeats, stage traces, matched budgets, clutter, and real transfer.}